# Deep Learning-Based Communication Over the Air

Sebastian Dörner, Sebastian Cammerer, Jakob Hoydis, and Stephan ten Brink

***Abstract*—End-to-end learning of communications systems is a fascinating novel concept that has so far only been validated by simulations for block-based transmissions. It allows learning of transmitter and receiver implementations as deep neural networks (NNs) that are optimized for an arbitrary differentiable end-to-end performance metric, e.g., block error rate (BLER). In this paper, we demonstrate that over-the-air transmissions are possible: We build, train, and run a complete communications system solely composed of NNs using unsynchronized off-the-shelf software-defined radios (SDRs) and open-source deep learning (DL) software libraries. We extend the existing ideas towards continuous data transmission which eases their current restriction to short block lengths but also entails the issue of receiver synchronization. We overcome this problem by introducing a frame synchronization module based on another NN. A comparison of the BLER performance of the "learned" system with that of a practical baseline shows competitive performance close to 1 dB, even without extensive hyperparameter tuning. We identify several practical challenges of training such a system over actual channels, in particular the missing channel gradient, and propose a two-step learning procedure based on the idea of transfer learning that circumvents this issue.**

## I. Introduction

The fundamental problem of communication is that of "reproducing at one point either exactly or approximately a message selected at another point" [1] or, in other words, reliably transmitting a message from a source to a destination over a channel by the use of a transmitter and a receiver. In order to come close to the theoretically optimal solution to this problem in practice, transmitter and receiver were subsequently divided into several processing blocks, each responsible for a specific sub-task, e.g., source coding, channel coding, modulation, and equalization. Although such an implementation is known to be sub-optimal [2], it has the advantage that each component can be individually analyzed and optimized, leading to the very efficient and stable systems that are available today.

The idea of deep learning (DL) communications systems [3], [4], on the contrary, goes back to the original definition of the communication problem and seeks to optimize transmitter and receiver jointly without any artificially introduced block structure. Although today's systems have been intensely optimized over the last decades and it seems difficult to compete with them performance-wise, we are attracted by the conceptual simplicity of a communications system that can learn to communicate over any type of channel without the need for prior mathematical modeling and analysis. The main contribution of this work is to demonstrate the practical potential and viability of such a system by developing a prototype consisting of two software-defined radios (SDRs) that learn to communicate over an actual wireless channel. To do so, we extend the ideas of [3], [4] to continuous data transmission, which requires dealing with synchronization issues and inter-symbol interference (ISI). This turns out to be a crucial step for over-the-air transmissions. Somewhat to our surprise, our implementation comes close to the performance of a well-designed conventional system.

S. Dörner, S. Cammerer, and S. ten Brink are with the Institute of Telecommunications, University of Stuttgart, Pfaffenwaldring 47, 70659 Stuttgart, Germany ({doerner,cammerer,tenbrink}@inue.uni-stuttgart.de).

J. Hoydis is with Nokia Bell Labs, Route de Villejust, 91620 Nozay, France (jakob.hoydis@nokia-bell-labs.com).

### A. Background

The fields of machine learning (ML) and, in particular, DL have seen very rapid growth during the last years; their applications extend now towards almost every industry and research domain [5], [6], [7]. Although researchers have tried to address communications-related problems with ML in the past (see [8], [9], [10] and references therein), it did not have any fundamental impact on the way we design and implement communications systems. The main reason for this is that extremely accurate and versatile system and channel models have been developed that enable algorithm design grounded in information theory, statistics, and signal processing, with reliable performance guarantees. Another reason is that, only since very recently, powerful DL software libraries (e.g., [11], [12], [13], [14], [15]), and specialized hardware, such as graphic processing units (GPUs) and field programmable gate arrays (FPGAs) (and increasingly other forms of specialized chips) are cheaply and widely available. These are key for training and inference of complex DL models needed for real-time signal processing applications. Thanks to these developments, several research groups have recently started investigations into DL applications in communications and signal processing using state-of-the-art tools and hardware. Among them are belief propagation for channel decoding [16], [17], one-shot channel decoding [18], compressed sensing [19], [20], coherent [21] and blind [22] detection for multiple-input multiple-output (MIMO) systems, detection algorithms for molecular communications [23], learning of encryption/decryption schemes for an eavesdropper channel [24], as well as compression [25]. The concept of learning an end-to-end communications system by interpreting it as an autoencoder ([26, Ch. 14]) was first introduced in [3] and [4]. Although a theoretically very appealing concept, it is not implementable in practice without modifications as the gradient of the channel is unknown at training time.

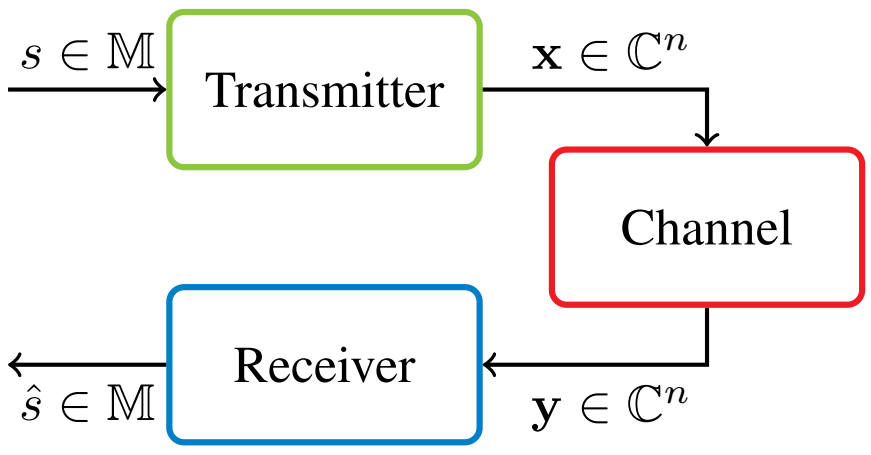

Fig. 1: Illustration of a simple communications system

The rest of this article is structured as follows: Section II describes the basic autoencoder concept and explains the challenges related to a hardware implementation. Section III contains a detailed description of our implementation, including two-phase training, channel modeling, and necessary extensions for continuous transmissions. Section IV presents performance results for a stochastic channel model as well as for over-the-air transmissions. Section V concludes the paper.

*Notations:* Boldface letters and upper-case letters denote column vectors and matrices, respectively. The $i$th element of vector $\mathbf{x}$ is denoted $x_i$. The notation $\mathbf{x}_i^j$ for $j \geq i$ denotes the vector consisting of elements $[x_i, x_{i+i}, \dots, x_j]^\mathsf{T}$ where $()^\mathsf{T}$ is the transpose operator. For a vector $\mathbf{x}$, $\arg\max(\mathbf{x})$ denotes the index (starting from 1) of the element with the largest absolute value. The operator $\mathrm{cs}\,[\mathbf{x}]_i$ denotes the circular shift of $\mathbf{x}$ by $i$ steps. $\mathbf{0}_n$ is the all-zero vector with $n$ dimensions. $\lceil\cdot\rceil$ and $\lfloor\cdot\rfloor$ denote the ceiling and floor functions, respectively.

## II. End-to-end Learning of Communications Systems

We consider a communications system consisting of a transmitter, a channel, and a receiver, as shown in Fig. 1. The transmitter wants to communicate a message $s \in \mathbb{M} = \{1, 2, \dots, M\}$ over the channel to the receiver. To do so, it is allowed to transmit $n$ complex baseband symbols, i.e., a vector $\mathbf{x} \in \mathbb{C}^n$ with a power constraint, e.g., $\|\mathbf{x}\|^2 \leq n$. At the receiver side, a noisy and possibly distorted version $\mathbf{y} \in \mathbb{C}^n$ of $\mathbf{x}$ can be observed. The task of the receiver is to produce the estimate $\hat{s}$ of the original message $s$. The block error rate (BLER) $P_e$ is then defined as

$$P_e = \frac{1}{M} \sum_s \Pr\left(\hat{s} \neq s | s\right). \tag{1}$$

Each message $s$ can be represented by a sequence of bits of length $k = \log_2(M)$, where the exact representation depends on the chosen labeling. Thus, the resulting communication rate is $R = k/n$ in $^{\text{bits}}/_{\text{channel use}}$. Note that, for now, an idealized communications system was assumed with perfect timing and both carrier-phase and frequency synchronization.

### A. The autoencoder concept

As explained in [4], the communications system described above can be interpreted as an autoencoder [26, Ch. 14]. This is schematically shown in Fig. 2. An autoencoder describes a deep neural network (NN) that is trained to reconstruct the input at the output. As the information must pass each layer, the network needs to find a robust representation of the input message at every layer.

The trainable transmitter part of the autoencoder consists of an embedding[1] followed by a feedforward NN (or multilayer perceptron (MLP)). Its $2n$-dimensional output is cast to an $n$-dimensional complex-valued vector by considering one half as the real part and the other half as the imaginary part (see [4]). Finally, a normalization layer ensures that the power constraint on the output $\mathbf{x}$ is met. The channel can be implemented as a set of layers with probabilistic and deterministic behavior, e.g., for an additive white Gaussian noise (AWGN) channel, Gaussian noise with fixed or random noise power $\sigma^2$ per complex symbol is added. Additionally, any other channel effect can be integrated, such as a tapped delay line (TDL) channel, carrier frequency offset (CFO), as well as timing and phase offset. The channel model used in our experiments is described in detail in Section III-B. The receiver consists of a transformation from complex to real values (by concatenating real and imaginary parts of the channel output), followed by a feedfoward NN whose last layer has a “softmax” activation (see [26]). Its output $\mathbf{b} \in (0,1)^M$ is a probability vector that assigns a probability to each of the possible messages. The estimate $\hat{s}$ of $s$ corresponds to the index of the largest element of $\mathbf{b}$. Throughout this work, we use feedforward NNs with dense layers and rectified linear unit (ReLU) activations, except for the last layer of the transmitter and receiver, which are linear and softmax, respectively.

The resulting autoencoder can be trained end-to-end using stochastic gradient descent (SGD). Since we have phrased the communication problem as a classification task, it is natural to use the cross-entropy loss function

$$L_{\text{loss}} = -\log\left(b_s\right) \tag{2}$$

where $b_s$ denotes the $s$th element of $\mathbf{b}$. As we deal with an autoencoder where the output should equal the input during training, we have a fixed number of $M$ different training labels. Note that the random nature of the channel acts as a form of regularization which makes is impossible for the NN to overfit, i.e., the receiver part never sees the same training example twice. Thus, an *infinite* amount of labeled training data is available by training with the same number of $M$ different labels over and over again. This is typically not the case in other ML domains. For more details about DL, we refer to the textbook [26]. After training, the autoencoder can be split into two parts (discarding the channel layers), the transmitter TX and the receiver RX, which define the mappings $\mathrm{TX}: \mathbb{M} \mapsto \mathbb{C}^n$ and $\mathrm{RX}: \mathbb{C}^n \mapsto \mathbb{M}$, respectively.

We want to emphasize a few important differences with respect to conventional quadrature amplitude modulation (QAM) systems. First, the learned constellation points at the output of the transmitter are not structured as in regular QAM. This is shown in Fig. 7 (cf. non-uniform constellations [27]). Second, the transmitted symbols are correlated over time since $\mathbf{x}$ forms $n$ complex symbols that represent a single message. This implies that a form of channel coding is inherently applied.

[1]An embedding is a function that takes an integer input $i$ and returns the $i$th column of a matrix, possibly filled with trainable weights. Alternatively, $s$ could be transformed into a “one-hot” vector before being fed into the NN.

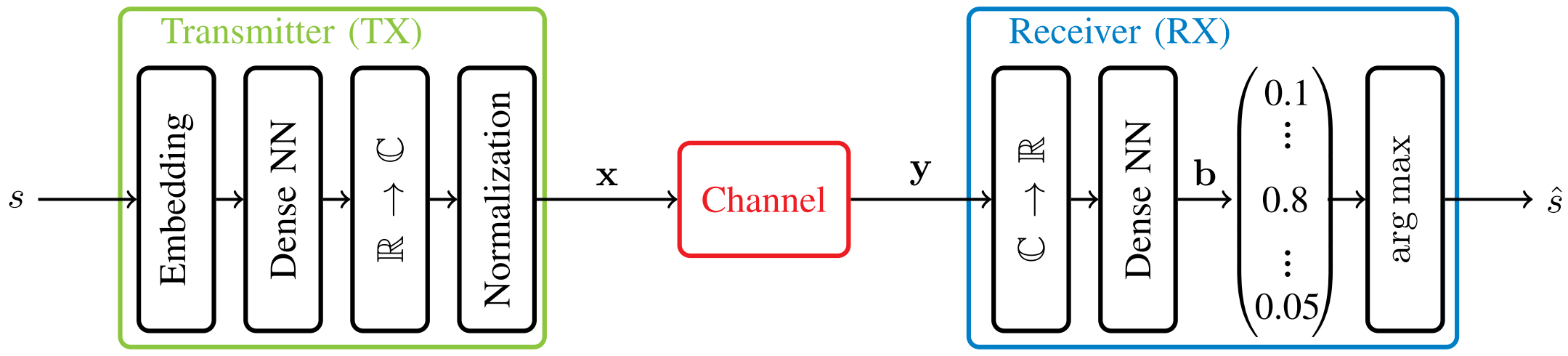

Fig. 2: Representation of a communications system as an autoencoder

### *B. Challenges related to hardware implementation*

The following practical challenges render the implementation of the autoencoder concept on real hardware difficult:

*1) Unknown channel transfer function:* The wireless channel (including some parts of the transceiver hardware, e.g., amplifiers, filters, analog-to-digital converter (ADC)) is essentially a black box whose output can be observed for a given input. However, its precise transfer function at a given time is unknown. This renders the gradient computation, which is crucially needed for SGD-based training, impossible. Thus, simple end-to-end learning of a communications systems is infeasible in practice. Instead, we propose a two-phase training strategy that circumvents this problem in Section III-A. The key idea is to train the autoencoder on a stochastic channel model and then finetune the receiver with training data obtained from over-the-air transmissions.

*2) Hardware effects:* The autoencoder model as described in Fig. 2 does not account for several important aspects that are relevant for a hardware implementation. First of all, a real system operates on samples and not symbols. Secondly, pulse shaping, quantization, as well as hardware imperfections such as CFO and timing/phase offset must be considered. Although all of these effects can be easily modeled, it requires a considerable amount of work to implement them as NN layers for a given DL library.

*3) Continuous transmissions:* The autoencoder of the last section cannot be trained for a large number of messages $M$ due to exponential training complexity. Already training for 100 bits of information (i.e., $M = 2^{100}$ possible messages)—which is small for any practical system—seems currently impossible. For this reason, the autoencoder must transmit a continuous sequence of a smaller number of possible messages. Unfortunately, this requires timing synchronization (to figure out which received samples out of a continuously recorded sample stream must be fed into the receiver NN) and makes the system vulnerable to sampling frequency offset (SFO). Moreover, pulse shaping at the transmitter creates ISI (over multiple messages) which must be integrated into the training process. We explain in Section III-C how we have changed the autoencoder architecture and training procedure to account for and combat these effects.

## III. Implementation

### *A. Two-phase training strategy*

The main advantage of describing the full end-to-end communications system as an autoencoder is that the training becomes straightforward. The gradients of all layers can be efficiently calculated by the backpropagation algorithm [28] which is implemented in all state-of-the-art DL libraries. However, as mentioned above, this concept only works when we have a differentiable mathematical expression of the channel's transfer function for each training example, as is the case for a stochastic channel model.

In order to overcome this issue, we adopt a two-phase training strategy based on the concept of transfer learning [29] that is visualized in Fig. 3. We first train the autoencoder using a stochastic channel model that should approximate as closely as possible the behavior of the expected channel (Phase I). When we deploy the trained transmitter and receiver on real hardware for over-the-air transmissions, the initial performance depends on the accuracy of this model. As there is generally a mismatch between the stochastic and the actual channel model (including transceiver hardware effects that have not been captured), communication performance suffers.

We compensate for this mismatch by *finetuning* the receiver part of the autoencoder (Phase II): The transmitter sends a large number of messages over the actual channel and the corresponding IQ-samples are recorded at the receiver. These samples, together with the corresponding message indices, are then used as a labeled data set for supervised finetuning of the receiver. The intuition behind this approach is as follows. For a good stochastic channel model, the transmitter learns representations $\mathbf{x}$ of the different messages $s$ that are robust with respect to the possible distortions created by the channel. Although the actual channel may behave slightly differently, the learned representations are good in the sense that we can train a receiver that is able to recover the sent messages with small probability of error. This idea is similar to what is widely done in computer vision. Deep convolutional NNs are trained on large data sets of images for a particular task, such as object recognition. In order to speed-up the training for other tasks, the lower layers are kept fixed while only the top layers are finetuned. The reasoning behind this is that the lower convolutional layers extract features of the images that are useful for many different tasks [26, Sec. 7.7].

### *B. Channel model for training*

As explained in the previous section, it is crucial to have a good stochastic channel model for initial end-to-end training of the autoencoder (Phase I). We have implemented such a channel model in Tensorflow [12] (including wrapper functions for Keras [13]) that comprises the follwing features (see Fig. 4):

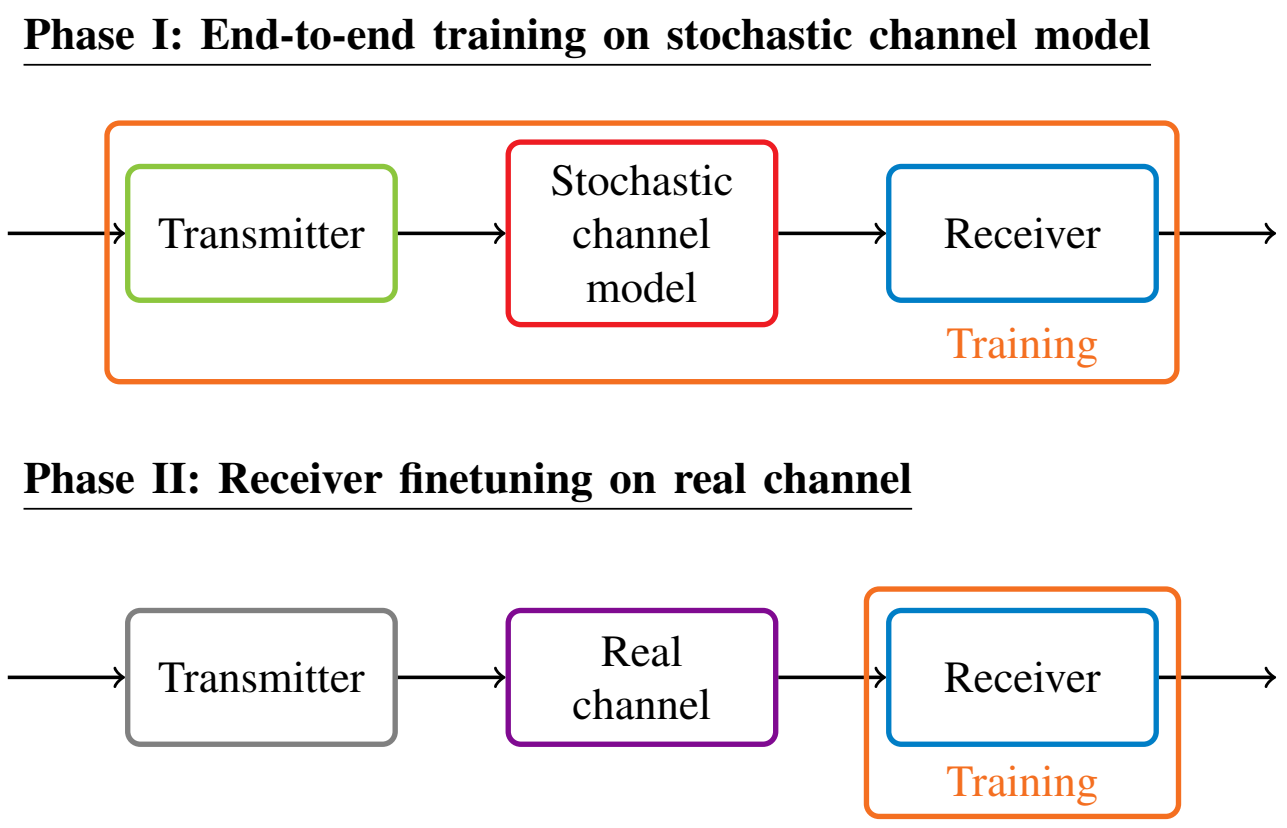

Fig. 3: Two-phase training strategy

- Upsampling & pulse shaping
- Constant sample time offset $\tau_{\text{off}}$
- Constant phase offset $\varphi_{\text{off}}$ and CFO
- AWGN

The channel model has no state, i.e., it generates a vector of received complex-valued IQ-samples for an input vector of complex-valued symbols, independently of previous or subsequent inputs.

*1) Upsampling & pulse shaping:* We define $\gamma \geq 1$ as the number of complex samples per symbol. First, the symbol input vector $\mathbf{x} \in \mathbb{C}^n$ is upsampled by inserting $\gamma - 1$ zeros after each symbol. Second, the resulting sample vector of size $N_{\text{msg}} = \gamma n$ is convolved with a discrete normalized root-raised cosine (RRC) filter[2] $g_{\text{rrc}}(t)$ with roll-off factor $\alpha$ and $L$ (odd) taps. The upsampled and filtered output $\mathbf{x}_{\text{rrc}} \in \mathbb{C}^{N_{\text{msg}}+L-1}$ has hence a length of $N_{\text{msg}} + L - 1$ complex-valued samples.

*2) Sample time offset:* Since the transmitter and receiver are not perfectly synchronized, there is a random offset of the sample time, denoted $\tau_{\text{off}}$. As the receiver is assumed to work directly on received IQ-samples without the help of any traditional synchronization algorithms, we can model this timing offset during the pulse-shaping step. This is done by randomly drawing a time offset $\tau_{\text{off}}$ from the interval $[-\tau_{\text{bound}}, \tau_{\text{bound}}]$[3] and computing the convolution of the upsampled input with the time-shifted filter $g_{\text{rrc}}(t - \tau_{\text{off}})$ as shown in Fig. 4. In practice, there will also be an SFO (which is related to the CFO because the same local oscillator is used). However, its effects are only visible over long time scales and do not need to be accounted for by our channel model as the frame synchronization procedure described in Section III-C4 will take care of this for continuous over-the-air transmissions.

*3) Phase offset & CFO:* Radio hardware suffers from a small mismatch between the oscillator frequency of the transmitter $f_{tx}$ and receiver $f_{rx}$, resulting in the CFO $f_{\text{cfo}} = f_{rx} - f_{tx}$. This leads to a rotation effect of the complex IQ-samples over time, which we model as

$$x_{\text{cfo},k} = x_{\text{rrc},k} e^{\mathrm{j}(2\pi k \Delta\varphi + \varphi_{\text{off}})} \tag{3}$$

[2]https://en.wikipedia.org/wiki/Root-raised-cosine_filter

[3]The frame synchronization procedure as described in Section III-C4 will ensure that the timing offset lies within this interval.

where $\Delta\varphi = \frac{f_{\text{cfo}}}{f_s}$ denotes the (normalized) inter-sample phase offset, $f_s$ is the sample frequency, and $\varphi_{\text{off}}$ is an additional unknown phase offset. The variables $\Delta\varphi \sim \mathcal{NC}(0, \sigma^2_{\text{CFO}})$ (truncated to the interval $[\varphi_{\text{min}}, \varphi_{\text{max}}]$) and $\varphi_{\text{off}} \sim \mathcal{U}(0, 2\pi)$ are randomly chosen for each channel input.

*4) AWGN:* Since we do not expect a multipath channel environment for our prototype, it is sufficient to consider an AWGN channel. The channel output $\mathbf{y} \in \mathbb{C}^{N_{\text{msg}}+L-1}$ is therefore modeled as

$$y_k = x_{\text{cfo},k} + n_k \tag{4}$$

where $n_k \sim \mathcal{CN}(0, \sigma^2)$. Fluctuations of the received signal strength due to propagation attenuation simply result in signal-to-noise ratio (SNR) variations while propagation-related phase rotations are already represented by the phase offset $\varphi_{\text{off}}$. If needed, it is straightforward to consider multipath fading in the stochastic channel model, as done in [4].

### C. Extensions for continuous transmissions

*1) Training over sequences of multiple messages:* Pulse shaping at the transmitter creates interference between the samples of adjacent symbols (ISI), where the number of samples that effectively interfere depends on the filter span $L$.[4] In traditional single-carrier communications systems, ISI is generally taken care of by matched filtering of the received samples with another RRC filter in combination with timing offset compensation (so that symbols are recovered without ISI by sampling at the symbol rate). The autoencoder, on the other hand, learns to handle ISI but does neither do matched filtering at the receiver side nor uses any existing form of synchronization

However, a problem arises if we train the autoencoder on the stochastic channel model for block-wise transmissions of individual messages and then use it for continuous transmission of sequences of multiple messages without a guard interval between them. The resulting inter-message interference (which the receiver has never seen during training) will severely degrade the communication performance. To avoid this problem, we assume during training that the transmitter (TX) generates samples corresponding to $2\ell + 1$ subsequent messages $\{s_{t-\ell}, \dots, s_t, \dots, s_{t+\ell}\}$ and let a sequence decoder (SD) decode message $s_t$ based on a slice $\mathbf{y}_{k_1}^{k_2}$ of the received signal vector $\mathbf{y} \in \mathbb{C}^{(2\ell+1)N_{\text{msg}}+L-1}$. This is shown in Fig. 5 which will be further explained in the next sections. In other words, the sequence decoder defines the mapping $\text{SD}: \mathbb{C}^{k_2-k_1+1} \mapsto \mathbb{M}$ (or batch-wise, meaning the decoding of $u$ sequences $\mathbb{C}^{k_2-k_1+1}$ in parallel, $\text{SD}: \mathbb{C}^{(k_2-k_1+1)\times u} \mapsto \mathbb{M}^u$) which is implemented as a combination of multiple NNs. The size and location of the slice $\mathbf{y}_{k_1}^{k_2}$ within the channel output (as determined by $k_1$ and $k_2$) as well as the number of padding messages $\ell$ are design parameters that are chosen depending on the length $L$ of the pulse shaping filter and the allowable complexity of the receiver. In the sequel, we will assume without loss of generality

$$k_1 = N_{\text{msg}} + \frac{L+1}{2}, \quad k_2 = 2\ell N_{\text{msg}} + \frac{L-1}{2} \tag{5}$$

[4]Also channels with memory create ISI which we do not consider here.

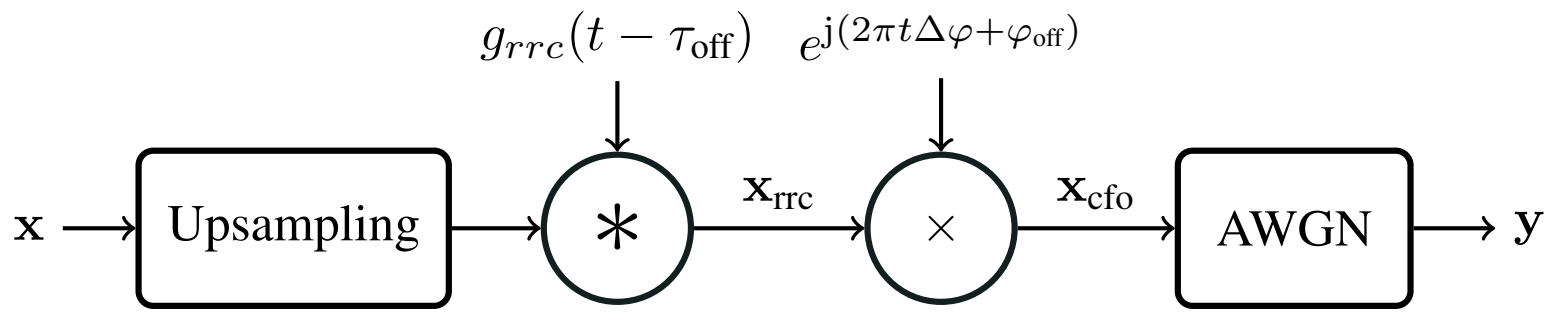

Fig. 4: Stochastic channel model for end-to-end training of the autoencoder

and define

$$N_{\text{seq}} = k_2 - k_1 + 1 = (2\ell - 1)N_{\text{msg}}. \tag{6}$$

Thus, the SD block is fed with $N_{\text{seq}}$ samples taken from the center of $\mathbf{y}$ corresponding to $2\ell - 1$ messages. The messages $s_{t-\ell}$ and $s_{t+\ell}$ are, thus, only used to mitigate edge effects.

*2) Phase offset estimation:* Although the autoencoder can learn to deal efficiently with random phase offsets as well as CFO, the performance for continuous transmissions can be improved by letting the receiver estimate the phase offset explicitly over multiple subsequent messages. We implement such a phase estimator (PE) as another dense NN that takes as input the slice of the channel output $\mathbf{y}_{k_1}^{k_2}$ and generates as output a single complex-valued scalar $h \in \mathbb{C}$ (with the corresponding transformations between real and complex domains where necessary). The received samples that are passed into the RX block for message detection (see Fig. 5) are multiplied by $h$ to compensate for the constant phase offset $\varphi_{\text{off}}$. This idea is based on the concept of radio transformer networks (RTNs) as explained in [30], [4] and is schematically shown in Fig. 5.

A couple of comments are in order: First, the phase estimator is integrated into the end-to-end learning process without consideration of any additional loss term in (2) for the estimated quantity $h$. Although it would be possible to consider the loss function $L_{\text{loss}} + w|h - e^{-\mathrm{j}\varphi_{\text{off}}}|^2$ for some $w > 0$ which adds the weighted mean squared error (MSE) to the cross-entropy, we have not seen any gains in doing so and set $w = 0$. Second, estimating $\varphi_{\text{off}}$ rather than $h \approx e^{-\mathrm{j}\varphi_{\text{off}}}$ did not lead to satisfactory results. A key problem is that $e^{-\mathrm{j}\varphi_{\text{off}}}$ is a periodic function, so that all estimates of $\varphi_{\text{off}}$ plus-minus multiples of $2\pi$ are equally good. This makes it difficult for gradient-based learning to converge to a stable solution. Third, we ignore CFO estimation and correction which could be implemented in a similar manner, e.g., another complex-valued scalar, say $r$, is estimated and the $k$th received sample is rotated by $r^k$. We have tried such an approach but did not observe any performance gains. This is likely because the considered IQ-sample sequence length in this work is relatively short so that explicit CFO compensation is not necessary and can be easily handled by the RX block. Fourth, the phase offset estimator could also be implemented as a recurrent neural network (RNN) that keeps an internal state from previous transmission sequences.[5] Thus, the estimation could be based on an essentially infinitely large observation horizon similar to a phase-locked loop (PLL) to further improve the performance. We have not experimented with such an architecture and keep it for future investigation.

[5]This requires the channel model to have an internal state as well.

*3) Feature extraction:* Similar to the phase estimator, the RX block could use the full (phase corrected) observation $h \cdot \mathbf{y}_{k_1}^{k_2}$ to estimate $s_t$. However, there are two problems with such an approach. First, it can happen that the observation window contains samples from several identical messages, e.g., $s_t = s_{t+3}$. This might be confusing for the receiver as it does not known which part of the input vector is responsible for the target output. Secondly, the larger the number of inputs, the larger the number of NN parameters which is undesirable from an implementation perspective. For this reason, we only feed the sub-slice $(\mathbf{y}_{k_1}^{k_2})_{l_1}^{l_2}$ into the RX block but concatenate it with a small number $F$ of features $\mathbf{f} \in \mathbb{C}^F$ that are extracted from $\mathbf{y}_{k_1}^{k_2}$ with the help of another dense NN (called "Feature Extractor (FE)" in Fig. 5). Our experiments have shown that even a small number of features, e.g., $F = 4$, significantly improves the performance. We assume for the rest of this work

$$l_1 = 1 + (\ell - 1)N_{\text{msg}} - \gamma, \quad l_2 = \ell N_{\text{msg}} + \gamma \tag{7}$$

and define

$$N_{\text{in}} = l_2 - l_1 + 1 + F = N_{\text{msg}} + 2\gamma + F. \tag{8}$$

With these definitions, the RX block decodes $s_t$ based on the $F$ features from the FE block concatenated with the $N_{\text{msg}}$ samples taken from the center of $\mathbf{y}$ to which $\gamma$ samples (representing one complex symbol) are pre- and appended.

Similar to the PE, the FE is integrated into the end-to-end training process without an explicit loss term in (2). It could be implemented as an RNN to potentially improve the performance, which is left to future work.

*4) Frame synchronization:* While CFO is not a big problem for the autoencoder—its block-wise decision process easily deals with a constant phase offset $\varphi_{\text{off}}$ while the overall phase shift due to CFO in a short IQ-sample sequence is very small—SFO is far more difficult to handle. SFO occurs when the oscillators in the transmitter and receiver run at slightly different frequencies, so that the derived sampling frequencies differ as well. Over a longer time period this causes the receiver to record more or less IQ-samples than have been sent. For instance, for an oscillator offset of 50 parts per million (ppm) between transmitter and receiver and a sample frequency of $f_s = 1\,\text{MHz}$, there are 50 more IQ-samples per second recorded at the receiver than sent by the transmitter. If enough IQ-samples are skipped or repeated, a whole IQ-symbol (or in the autoencoder case a whole message) can be created or lost. This effect leads to the very little understood concept of the *insertion and deletion* channel, for which not even the capacity is known [31].

In traditional communications systems, SFO is commonly handled by tracking of the actual timing offset through a PLL and resampling of the received signal at the desired target time instances (see, e.g., [32]). For our implementation, we have not adopted this approach and developed a frame synchronization module based on another NN that works as follows (see Fig. 6).

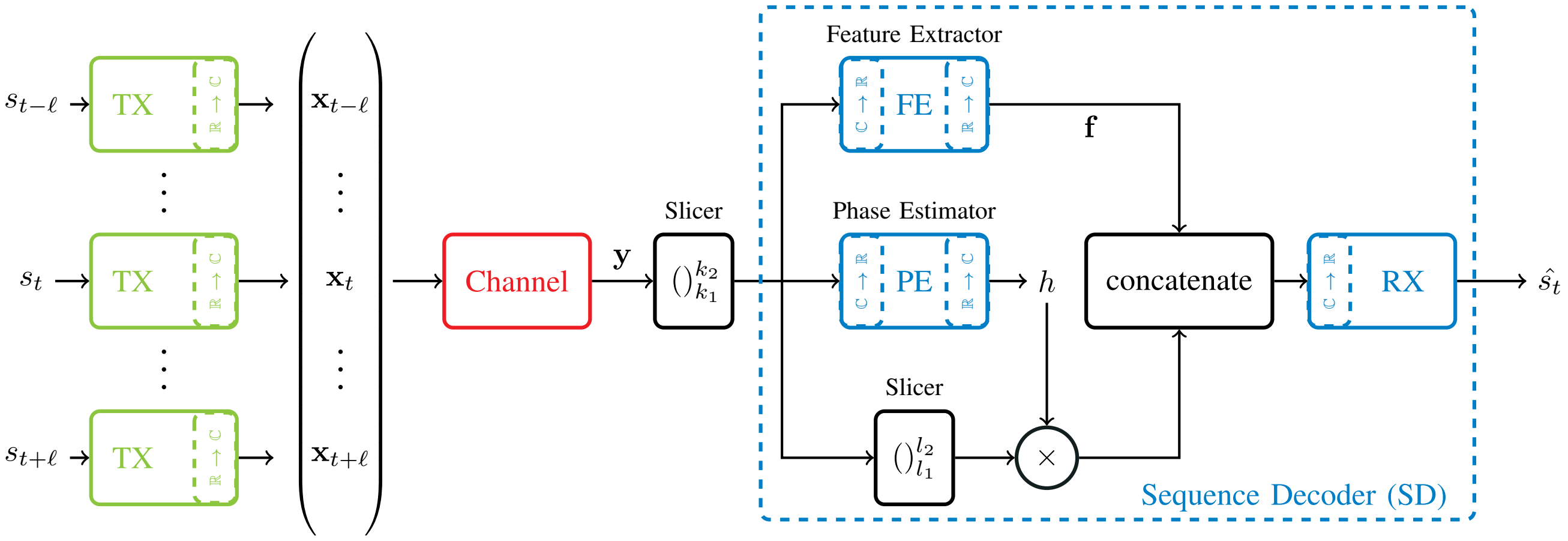

Fig. 5: Flow graph of the sequence-based end-to-end training process including phase offset estimation and feature extraction

Assume that the receiver observes the infinite sequence of IQ-samples $y_1, y_2, \ldots$. The goal of the frame synchronization module is to estimate the sample offset $\hat{\imath}$ for a "frame" of $B$ messages, spanning $N = BN_{\text{msg}}$ samples. In other words, for a frame $\mathbf{y}_k^{k+N-1}$ starting at sample index $k$, the frame synchronization module will estimate by how many samples $\hat{\imath}$ the frame must be shifted to the left or right such that it contains precisely $B$ messages. The resulting frame $\mathbf{y}_{k+\hat{\imath}}^{k+\hat{\imath}+N-1}$ will then be rearranged into a batch of sample sequences of length $N_{\text{seq}}$ that are fed into the SD block for detection of the individual messages. Before we proceed with the description of the overall frame synchronization and detection process, we require the following definitions. Denote by $\mathbf{\Pi}(\mathbf{v}, d, q)$ the function that returns for an $r$-dimensional vector $\mathbf{v}$ and positive integers $d, q$, with $1 \leq d \leq r$ and $m = 1 + \lfloor (r-d)/q \rfloor$, the $d \times m$ matrix

$$\mathbf{\Pi}\left(\mathbf{v}, d, q\right) = \begin{bmatrix} v_1 & v_{1+q} & \cdots & v_{1+(m-1)q} \\ v_2 & v_{2+q} & \cdots & v_{2+(m-1)q} \\ \vdots & \vdots & \ldots & \vdots \\ v_d & v_{d+q} & \cdots & v_{d+(m-1)q} \end{bmatrix}. \tag{9}$$

The sample offset $\hat{\imath}$ for the frame $\mathbf{y}_k^{k+N-1}$ is estimated by Algorithm 1 with input $\mathbf{\Pi}\left(\mathbf{y}_k^{k+N-1}, N_{\text{seq}}, 1\right)$. This algorithm computes for each of the $m = 1 + N - N_{\text{seq}}$ possible sub-sequences of length $N_{\text{seq}}$ of $\mathbf{y}_k^{k+N-1}$ a probability vector over the possible sample offsets $\boldsymbol{\tau} \in (0,1)^{N_{\text{msg}}}$, averages them[6], and picks the most likely offset as estimate for the entire frame. The vector $\boldsymbol{\tau}$ is computed by an offset estimator (OE), which defines the mapping $\text{OE} : \mathbb{C}^{N_{\text{seq}}} \mapsto (0,1)^{N_{\text{msg}}}$ (or batch-wise $\text{OE} : \mathbb{C}^{N_{\text{seq}} \times u} \mapsto (0,1)^{N_{\text{msg}} \times u}$) and is implemented as a dense NN with $N_{\text{msg}}$ outputs and softmax output activation. The value of $\tau = \arg\max(\boldsymbol{\tau})$ lies withing the range $[1, N_{\text{msg}}]$ which is mapped to the offset $\hat{\imath}$ through the function $\tau - 1 - N_{\text{msg}} \left\lfloor \frac{\tau - 1}{\lceil N_{\text{msg}}/2 \rceil} \right\rfloor$. Thus, $\hat{\imath} \in [-\lfloor N_{\text{msg}}/2 \rfloor, \lfloor (N_{\text{msg}}-1)/2 \rfloor]$. For example, for $N_{\text{msg}} = 16$, $\tau \in [1, 16]$ is mapped to $\hat{\imath} \in [-8, 7]$. The estimation of the individual sample offsets for each column of the input $[\boldsymbol{\tau}_1, \ldots, \boldsymbol{\tau}_m]$ in Algorithm 1 can be computed in parallel by most DL libraries. It is also possible to average $\boldsymbol{\tau}$ over every $q$th possible sub-sequence to decrease the complexity. The training of the OE could be integrated into the end-to-end training process of the autoencoder. However, we have not done this here and trained it afterwards in an isolated fashion, where we used the output signals of the already trained transmitter part of the autoencoder to generate the input training data for the OE.

[6]Note that we apply a circular shift to the probability vectors before they are added corresponding to their relative position $u$ within the frame.

**Algorithm 1:** EstimateOffset

**Input :** Batch of $m$ IQ-sample sequences $\mathbf{Y} = [\mathbf{y}_1, \ldots, \mathbf{y}_m] \in \mathbb{C}^{N_{\text{seq}} \times m}$
**Output:** Sample offset $\hat{\imath}$

```
/* Initialization                                  */
```
$\boldsymbol{\tau} \leftarrow \mathbf{0}_{N_{\text{msg}}}$
```
/* Batch-estimate sequence offsets                 */
```
$[\boldsymbol{\tau}_1, \ldots, \boldsymbol{\tau}_m] \leftarrow \text{OE}\left(\mathbf{Y}\right)$
```
/* Calculate overall batch offset                  */
```
**for** $u \leftarrow 1$ **to** $m$ **do**
  $\boldsymbol{\tau} \leftarrow \boldsymbol{\tau} + \frac{1}{m} \text{cs}\left[\boldsymbol{\tau}_u\right]_{u-1}$
**end**
```
/* Compute offset and center                       */
```
$\tau = \arg\max\left(\boldsymbol{\tau}\right)$
$\hat{\imath} \leftarrow \tau - 1 - N_{\text{msg}} \left\lfloor \frac{\tau - 1}{\lceil N_{\text{msg}}/2 \rceil} \right\rfloor$

*5) Full decoding algorithm:* With the extensions of the autoencoder explained above, we are now ready to define the full decoding algorithm that we have used in our setup. The correspoding pseudo-code is provided in Algorithm 2. For the infinite sequence of IQ-samples $y_1, y_2, \ldots$, the decoding algorithm will find the start index of the first frame using Algorithm 1 and decode all messages in this frame using

Algorithm 3, before jumping to the next frame whose offset is computed by Algorithm 1 and so on. Since the SD requires a sample sequence of length $N_{\text{seq}} = (2\ell - 1)N_{\text{msg}}$ to decode a single message, the first and last $\ell - 1$ messages in each frame cannot be decoded. For this reason, the start index of frame $b > 1$ is chosen such that the last $\ell - 1$ messages of frame $b - 1$ are the first decodable messages of frame $b$. The very first $\ell - 1$ messages are hence never decoded.

While the sample offset of the first frame $\hat{\imath}_1$ can take any possible value, we chose the frame length $N$ such that the offset of frame two $\hat{\imath}_2$ (and of all subsequent frames) can be only one of $\{-1, 0, 1\}$. For example, with an SFO of $50\,\text{ppm}$ and a sample frequency of $1\,\text{MHz}$, one expects to loose/gain one sample every $20{,}000$ samples. For any $N$ smaller than this value, the aforementioned condition holds and can even be integrated into Algorithm 3. That is what we have done in our implementation. Thus, by skipping or repeating one sample, the OE can adjust $\tau_{\text{off}}$ seen by the SD by exactly plus-minus one $T_{\text{s}} = f_s^{-1}$. As a result, the minimum width of the sample time offset range $[-\tau_{\text{bound}}, \tau_{\text{bound}}]$ for which the autoencoder must be trained over the stochastic channel model (see III-B) is $T_{\text{s}}$. This defines $\tau_{\text{bound,min}} = T_{\text{s}}/2$. Note that this is only the absolut minimum value for $\tau_{\text{bound}}$ given that the OE's sample offset decision is always $100\,\%$ correct. But since we do not want the SD to fail because of a slightly wrong OE decision, we train the SD for a wider range of sample time offset by setting $\tau_{\text{bound}} = T_{\text{s}}$.

**Algorithm 2:** DecodeSequence

```
Input  : IQ-samples y_1, y_2, ...
         Number of samples per frame N
         Number of inputs for SD N_seq
         Number of samples per message N_msg
Output: Sequence of decoded frames ŝ_1, ŝ_2, ...

/* Initialization                              */
i ← 1, b ← 1

/* Frame-by-frame sequence decoding            */
repeat
    î_b ← EstimateOffset(Π(y_i^{i+N-1}, N_seq, 1))
    i_b ← i + î_b
    ŝ_b ← DecodeFrame(Π(y_{i_b}^{i_b+N-1}, N_seq, N_msg))
    i ← i_b + N − 2(ℓ − 1)N_msg
    b ← b + 1
```

**Algorithm 3:** DecodeFrame

```
Input  : Batch of B − 2(ℓ − 1) IQ-sample sequences
         Y = [y_1, ..., y_{B−2(ℓ−1)}] ∈ C^{N_seq × B−2(ℓ−1)}
Output: Decoded frame ŝ = [ŝ_1, ..., ŝ_{B−2(ℓ−1)}]^T

/* Batch-decode all sequences                  */
ŝ ← SD(Y)
```

## IV. RESULTS

In Sections IV-B and IV-C respectively, we evaluate the performance of the trained end-to-end communications system via simulations and measurements. In both cases, we use the same set of system parameters as described in Table I. The NN layouts of the transmitter (TX), receiver (RX), feature extractor (FE), phase offset (PE) and offset estimator (OE) are provided in Table II. PE, FE, and OE operate on sequences of $2\ell - 1 = 11$ subsequent messages (i.e., $\ell = 6$). With $n = 4$ complex symbols per message and $\gamma = 4$ complex samples per symbol, we have $N_{\text{msg}} = 16$ and $N_{\text{seq}} = 11 \times 4 \times 4 = 176$ $(352)$ complex (real) values. The FE block generates $F = 4$ complex-valued features $\mathbf{f}$, so that the input to the RX block consists of $N_{\text{in}} = 16{+}8{+}4 = 28$ $(56)$ complex (real) values. A frame consists of $B = 100$ messages, i.e., $N = 1600$ samples. This ensures that the sample offsets $\hat{\imath}_b$ for $b > 1$ can only take values from $\{-1, 0, 1\}$ even for severe SFO.

The autoencoder is trained over the stochastic channel model using SGD with the Adam optimizer [33] at learning rate $0.001$ and constant $E_b/N_0 = 9\,\text{dB}$. We have trained over 60 epochs of $5{,}000{,}000$ random messages with an increasing batch size after every 10 epochs, taking the values $\{50, 100, 500, 1000, 5000, 10{,}000\}$. The sample offset estimator (OE) was trained in a similar manner with larger max/min sample time offsets, namely $\tau_{\text{bound}} = 8\,T_s$. We have not carried out any extensive hyperparameter optimization (over number/size of layers, activation functions, training $E_b/N_0$, etc.), and simply tried a few different NN architectures from which the best performing is shown in Table II. Training of the autoencoder took about one day on an NVIDIA TITAN X (Pascal) consumer class GPU.

Finetuning of the receiver on each of the tested channels was done with 20 sequences of $1{,}600{,}000$ received IQ-samples to account for different initial timing and phase offsets. To gather the training data from sequences we demodulated them with the initially trained OE and SD system first. The OE plays a crucial role in this process, because the training labels are sliced out of the recorded sequences according to its offset estimations. Since we know the originally transmitted messages, which are used as targets, we can then label the corresponding sliced-out IQ-sample sequences of length $N_{seq}$, which are used as inputs. We only used sequences where the initial receiver had a BLER performance between $10^{-2}$ and $10^{-4}$ as finetuning training data. We experienced that finetuning with sequences for which the initial receiver had a performance worse than about $10^{-2}$ resulted in performance improvements in the low SNR range. However, it also led to a high error floor in the high SNR range. Since finetuning with very noisy sequences obscures important structure in the data, the receiver looses its ability to generalize to high SNR. We also observed a marginal gain by finetuning with sequences for which the BLER was below $10^{-4}$. Finetuning was done over one epoch per batch size, with the batch size taking the values $\{1000, 5000, 10{,}000\}$. This process is very fast and took less than a minute. We only trained for such a short number of iterations because the NN started to overfit. Further improvement of this finetuning process is one of our goals for the future. The OE was not finetuned as it already achieved

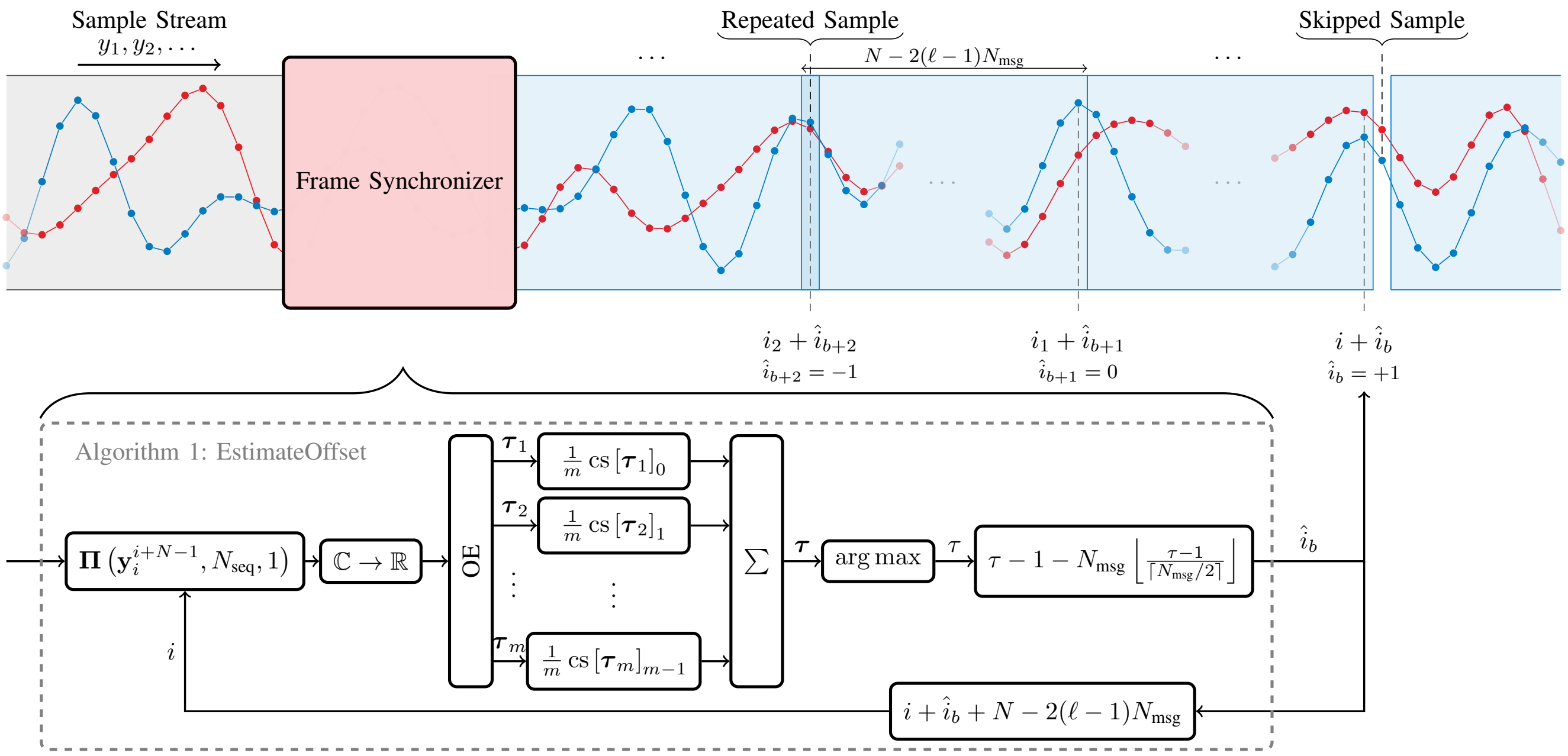

Fig. 6: Illustration of the frame synchronization module for a continuous sample sequence input $y_1, y_2, \ldots$

satisfactory performance.[7]

As a baseline system for comparison, we use a GNU Radio (GNR) differential quadrature phase-shift keying (DQPSK) transceiver, relying on polyphase filterbank clock recovery [32] with 32 phase filters (each with 1408 RRC taps) and a normalized loop bandwidth of 0.002513, which is based on the performance one could realize in hardware with the PLL of the used SDR. We leveraged the GNR DQPSK Mod and GNR DQPSK Demod Modules which are already implemented and documented in the official GNR software package. Since the rate loss due to differential modulation is negligible for long sequences, this system has the same rate of $R = 2$ bits/channel use and also does not need any kind of pilot signaling (except the initial symbol). For a better performance over our USRP testbed, we have imposed a power constraint on the maximum absolute value of each complex symbol generated by the autoencoder, i.e., $\max_i |x_i| \leq 1$, which is more strict than $\|\mathbf{x}\|^2 \leq n$. We observed that the autoencoder achieves a better performance on the stochastic channel when the normalization is set to an average power constraint, but, for reasons under current investigation, this did not translate into performance gains on real hardware and is still part of our current and future work. For a fair comparison with the baseline we have then scaled the learned constellation to have the same average power as DQPSK.

### A. *Learned Constellations*

As the weigths of the transmitter part of the autoencoder are fixed after the initial training on the stochastic channel, all possible output signals can be generated. Fig. 7 shows a scatter plot of the learned IQ-constellations at the transmitter. Recall that one message is composed of $n = 4$ complex symbols and that there are $M = 256$ different messages in total. We can make a couple of interesting observations from this figure. First, most symbols lie on the unit circle which shows that the autoencoder has learned to make efficient use of the available energy per message. Second, while the last three symbols of all messages are uniformly distributed on the unit circle or close to zero, the first symbols are all located in the second quadrant. This indicates that they are used, implicitly, for phase offset estimation. Third, some messages have a symbol close to zero, hinting at a form of time sharing between the symbols of different messages.

[7]Finetuning of the OE can be done by transmitting known synchronization sequences, such as Barker codes, in fixed intervals. These can be detected by correlation to compute timing offset labels that are used as ground truth for received sample sequences for which the timing offset is not known a priori.

TABLE I: Summary of system parameters

| Parameter | Symbol | Value |
|---|---|---|
| Bandwidth | $W$ | 500 kHz |
| Sequence length of one message | $N_{\text{msg}}$ | 16 samples (4 symbols) |
| Carrier frequency | $f_c$ | 2.35 GHz |
| Frame size | $B$ | 100 messages |
| Sample frequency | $f_s$ | 2 MHz |
| Sample duration | $T_s = f_s^{-1}$ | 0.5 $\mu$s |
| Symbol duration | $T_{\text{sym}}$ | 2 $\mu$s |
| Upsampling factor | $\gamma$ | 4 |
| Roll-off factor | $\alpha$ | 0.35 |
| Filter span (stochastic model) | $L$ | 31 samples |
| Number of messages | $M$ | 256 |
| Complex symbols per message | $n$ | 4 |
| Communication rate | $R$ | 2 bits/channel use |

TABLE II: Layout of all used NNs

| Transmitter (TX) : | Parameters | Output dimensions |
|---|---|---|
| Input | 0 | 1 (integer $s \in \mathbb{M}$) |
| Embedding | 65,536 | 256 |
| Dense (relu) | 65,792 | 256 |
| Dense (relu) | 2056 | 8 |
| Normalization | 0 | 8 ($n = 4$ complex symbols) |
| **Receiver (RX) :** | | |
| Input | 0 | 56 (28 complex samples) |
| Dense (relu) | 14,592 | 256 |
| Dense (relu) | 65,792 | 256 |
| Dense (softmax) | 65,792 | $M = 256$ |
| **PE and FE:** | | |
| Input | 0 | 352 (176 complex samples) |
| Dense (relu) | 90,368 | 256 |
| PE: Dense (linear) | 514 | 2 (1 complex symbols) |
| FE: Dense (linear) | 2056 | 8 (4 complex symbols) |
| **Offset estimator (OE):** | | |
| Input | 0 | 352 (176 complex samples) |
| Dense (relu) | 90,368 | 256 |
| Dense (relu) | 65,792 | 256 |
| Dense (relu) | 65,792 | 256 |
| Dense (softmax) | 4112 | 16 |

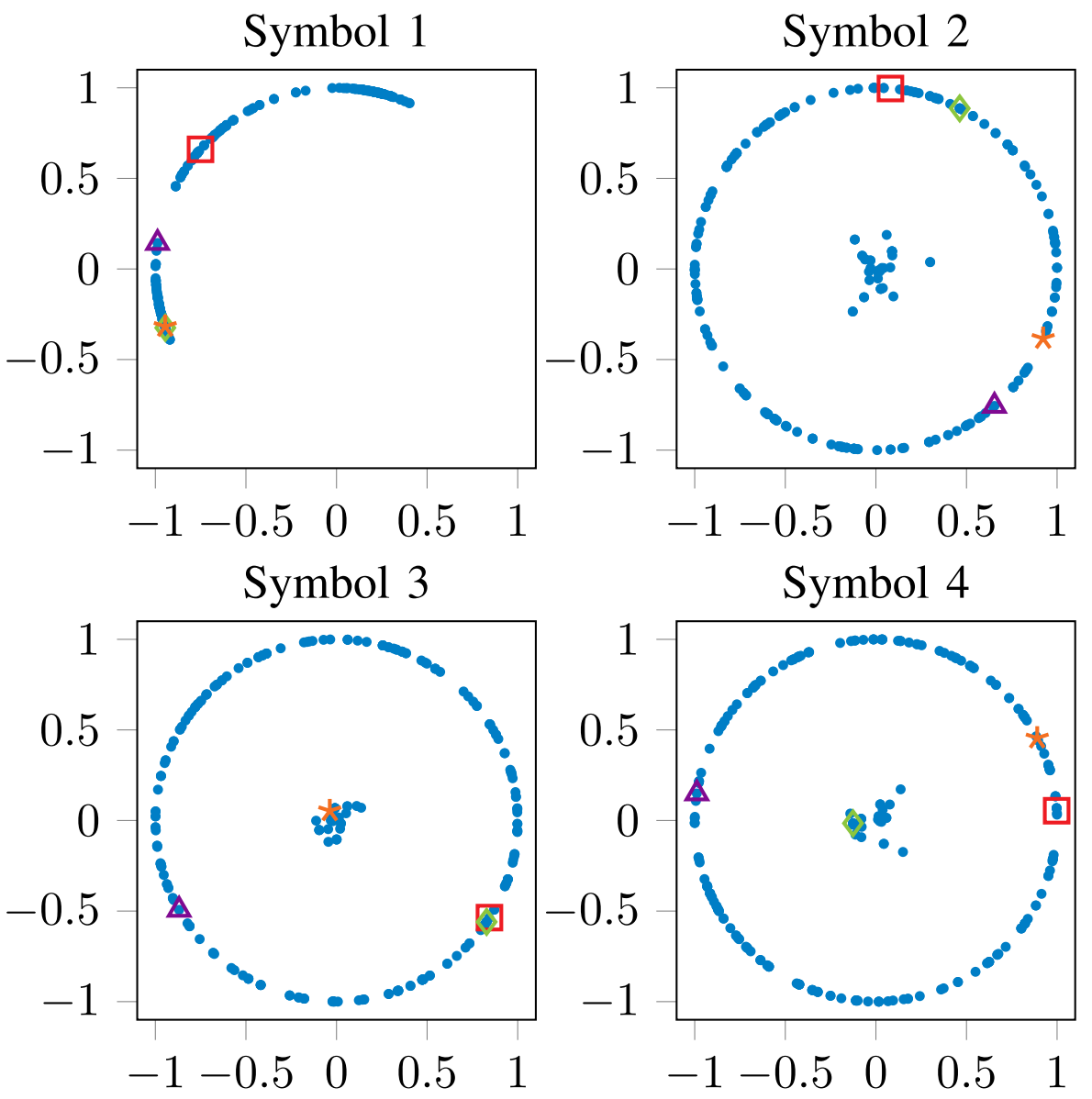

Fig. 7: Scatter plot of the learned constellations for all $M = 256$ messages. The symbols of four individual messages are highlighted by different markers.

As an example, Fig. 8a shows the real part $\text{Re}(\mathbf{x})$ of a sequence created by the autoencoder for the input message sequence $\{12, 224, 103, 177, 210, 214, 227, 88, 239, 100, 4\}$ and a random DQPSK signal of the same length. We observe less *structure* for the autoencoder sequence, i.e., less periodicity in the signal when compared to the DQPSK sequence which is a direct result of the non-uniform constellation. Fig. 8b shows the real part of the sequences $\text{Re}(\mathbf{y})$ recorded at the receiver for the transmitted sequences of Fig. 8a. Both sequences shown in Fig. 8b were recorded over-the-air (see IV-C) at a transmit gain of 65 dB and show the effects of a random phase offset, random timing offset and added noise. The received autoencoder sequence shown in Fig. 8b is a typical input sample sequence of length $N_{\text{seq}} = 176$ for the SD, which has to decode it to message $s = 214$. We provide this plot to illustrate the shape of signals from which the sequence decoder is able to detect messages. The imaginary part is omitted as it does not provide any additional insight.

TABLE III: Parameters of the GNR channel model with dynamic SFO and CFO

| Parameter | Value |
|---|---|
| Oscillator Accuracy | 20 parts per million |
| SFO Standard Deviation | 0.01 Hz/sample |
| Maximum SFO | 40 Hz |
| CFO Standard Deviation | 11.75 Hz/sample |
| Maximum CFO | 47 kHz |

### B. *Performance over simulated channels*

As a first benchmark, we have tested the performance of the autoencoder over a GNR channel model[8] that accounts for some of the realistic impairments we expect to see for over-the-air transmissions. To do so, we have used the transmitter to generate 15 sequences of 100,000 transmitted messages per transmit gain step which are individually fed into a chain of several GNR blocks simulating the radio transmissions. These blocks comprise, an RRC filter with resampling rate $\gamma = 4$ (Polyphase Arbitrary Resampler module, the same as used by the GNR DQPSK Mod module), SFO and CFO channel models followed by addition of a Gaussian noise source. The parameters of these blocks are listed in Table III. We refer for further details on how these blocks work and the precise meaning of these parameters to the GNR documentation.[9] The first five output sequences of this channel model are used for finetuning of the receiver system consisting of RX, PE, and FE, while the remaining 10 sequences are used to evaluate the BLER performance.

Fig. 9 shows BLER curves for the autoencoder (with and without finetuning) and the GNR DQPSK baseline. For reference, we also added a theoretical curve for DQPSK over an AWGN channel without CFO and SFO. The performance of the autoencoder is about 1 dB worse than that of the GNR DQPSK baseline over the full range of $E_b/N_0$. Thus, although training was done for a fixed $E_b/N_0 = 9\,\text{dB}$, the autoencoder is able to generalize to a wide range of different $E_b/N_0$ values. The figure also shows that the gain of finetuning is very small because the mismatch between the stochastic and the GNR channel model is small. We suspect that the performance gap to the baseline is due to differences in CFO modeling and could be made smaller by changes to the stochastic channel model.

### C. *Performance over real channels*

As schematically shown in Fig. 10, our testbed for over-the-air transmissions consisted of a USRP B200 as transmitter and a USRP B210 as receiver. Both were connected to two computers which communicated via Ethernet with a server equipped with an NVIDIA TITAN X (Pascal) consumer class GPU which was used for initial training, finetuning, and BLER computations. The USRPs were located indoors in a hallway with a distance of 46 m, having an unobstructed line-of-sight (LOS) path between them. We did not change their positions during data transmissions. As our implementation is not yet ready for real-time operation (we are currently decoding with

[8] https://gnuradio.org/doc/sphinx-3.7.2/channels/channels.html
[9] https://gnuradio.org/doc/doxygen/index.html

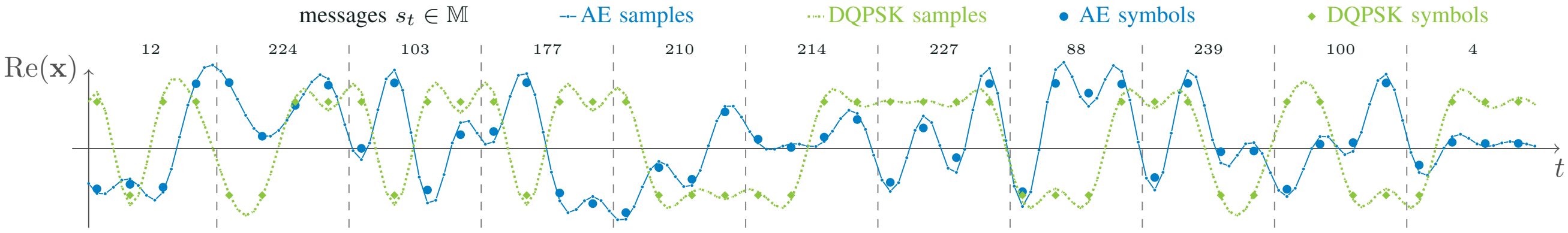

(a) Real part of the transmitted RRC pulse-shaped sample sequences and the original constellation symbols

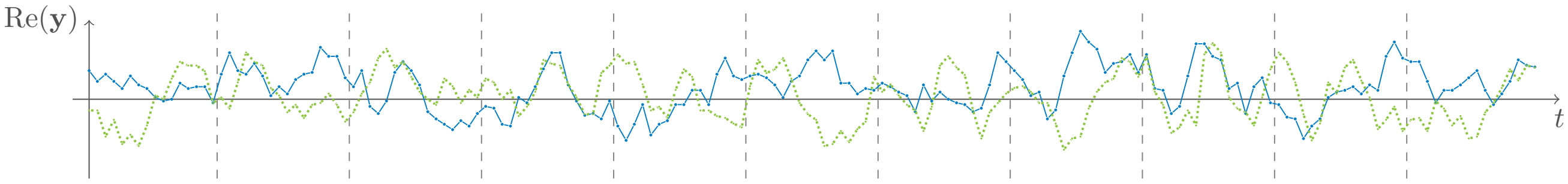

(b) Real part of the received sample sequences, recorded over-the-air at a transmit gain of 65 dB

Fig. 8: Sample and symbol sequences of the autoencoder (blue), for corresponding messages $s_t$, and DQPSK (green)

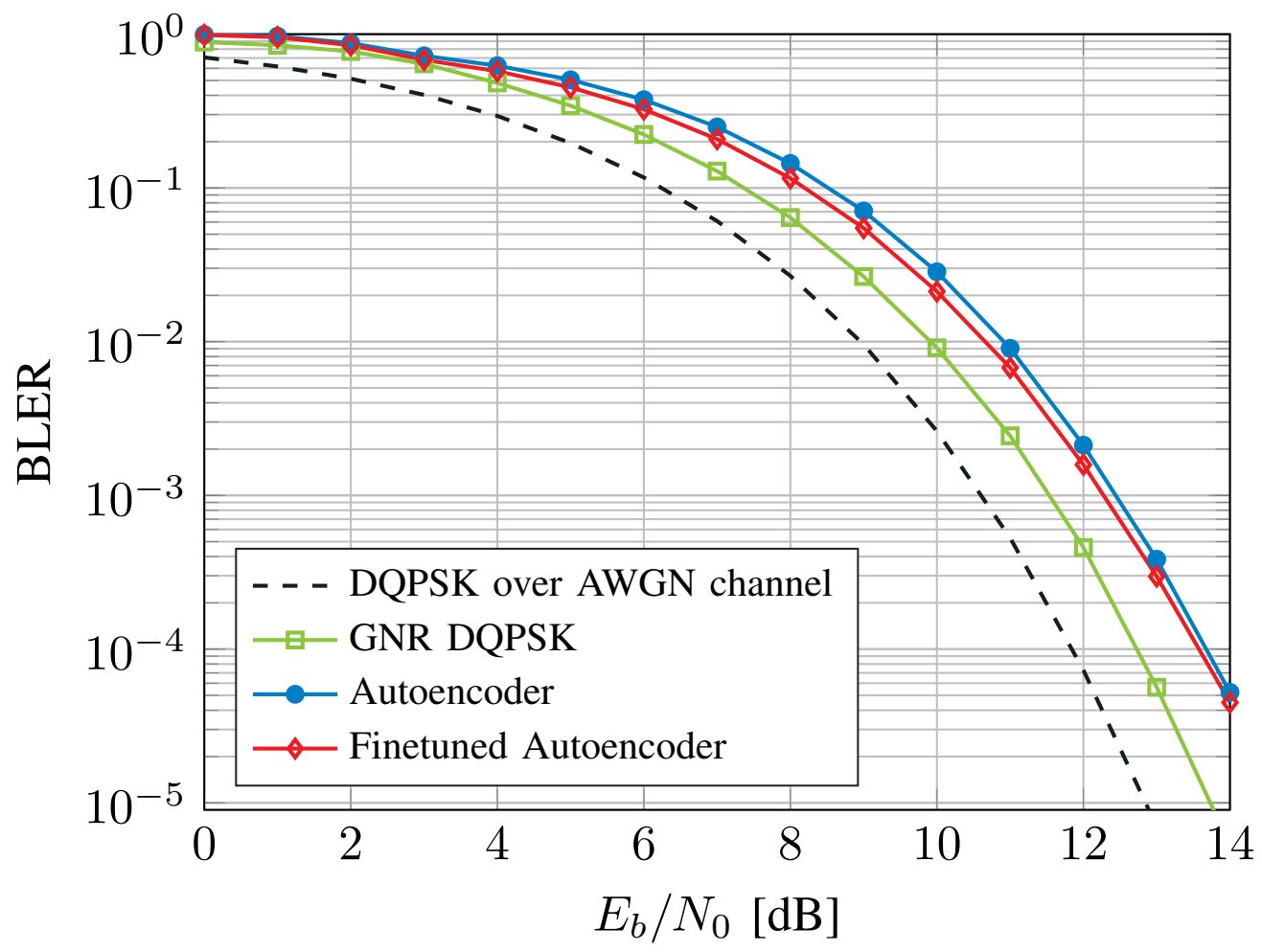

Fig. 9: Simulated BLER over the GNR channel model

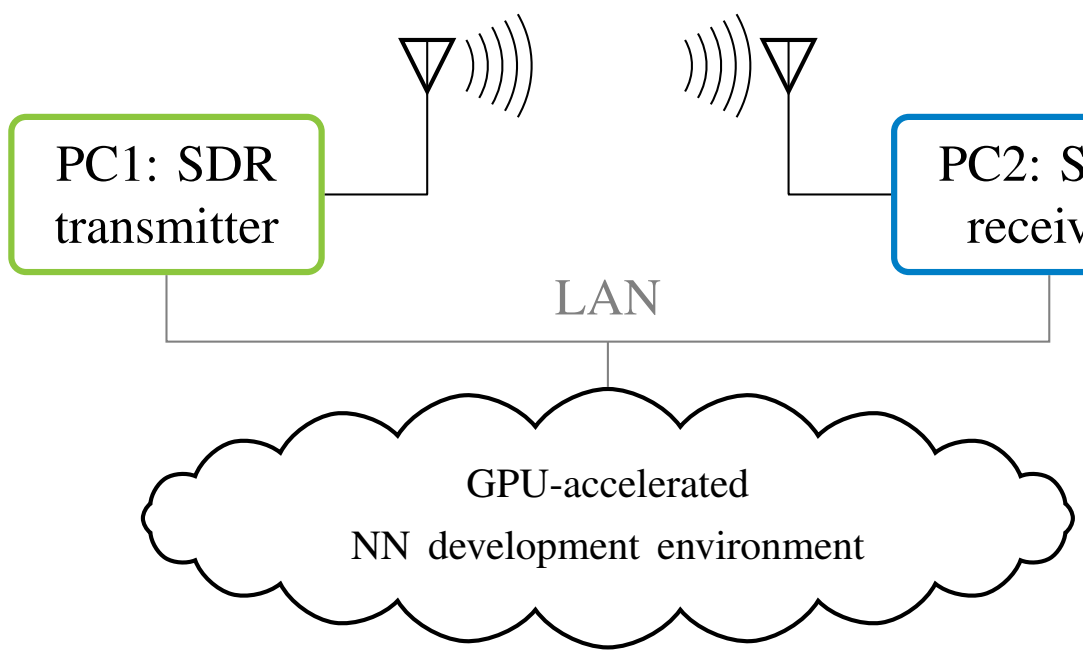

Fig. 10: Overview of the SDR testbed

about 10 kbit/s), we recorded 15 random transmission sequences per gain step consisting of 1,600,000 payload samples plus about one second of recording before and after payload transmission (which adds another $2 \times 2{,}000{,}000$ samples at sample rate 2 MHz) for 15 different USRP transmit gains. These were then processed offline, the payload was detected by pilots within the decoded message sequence and the BLER was averaged over the sequences for each gain.

Fig. 11a shows the over-the-air BLER for the autoencoder and the GNR DQPSK baseline versus the transmit gain. Without finetuning, there is a gap of 2 dB between the autoencoder and the baseline at a BLER of $10^{-4}$. This gap can be reduced to 1 dB through finetuning. The increased importance of finetuning in this scenario is due to a greater mismatch between the stochastic channel model used for training and the actual channel. Fig. 11b shows the same set of curves for transmissions over a coaxial cable from which similar observations can be made.

## V. Conclusions and Outlook

We have demonstrated that it is possible to build a point-to-point communications system whose entire physical layer processing is carried out by NNs. Although such a system can be rather easily trained as an autoencoder for any stochastic channel model, substantial effort is required before it can be used for over-the-air transmissions. The main difficulties we needed to overcome were the difference between the actual channel and the channel model used for training, as well as time synchronization, CFO, and SFO. The BLER performance of our prototype is about 1 dB worse than that of a GNR DQPSK baseline system. We expect, however, that this gap can be made significantly smaller (and maybe even turned into a gain) by hyperparameter tuning and changes to the NN architectures. We are currently building an implementation of our prototype as GNR blocks.

As this is the first prototype of its kind, our work has raised many more questions than it has provided answers. Most of the former can already be found in [4]. A key question is how to train such a system for an arbitrary channel for which no channel model is known. Moreover, it is crucial to enable on-the-fly finetuning to make the system adaptive to varying channel conditions for which it has not been initially trained. This could be either achieved through sporadic transmission of known messages or through an outer error-correcting code that would allow gathering a finetuning dataset on-the-fly.

## Acknowledgments

We would like to thank Maximilian Arnold and Marc Gauger for many helpful discussions and their contributions to a powerful baseline system (as well as for debugging the SDRs). We also acknowledge support from NVIDIA through their academic program which granted us a TITAN X (Pascal) graphics card that was used throughout this work.

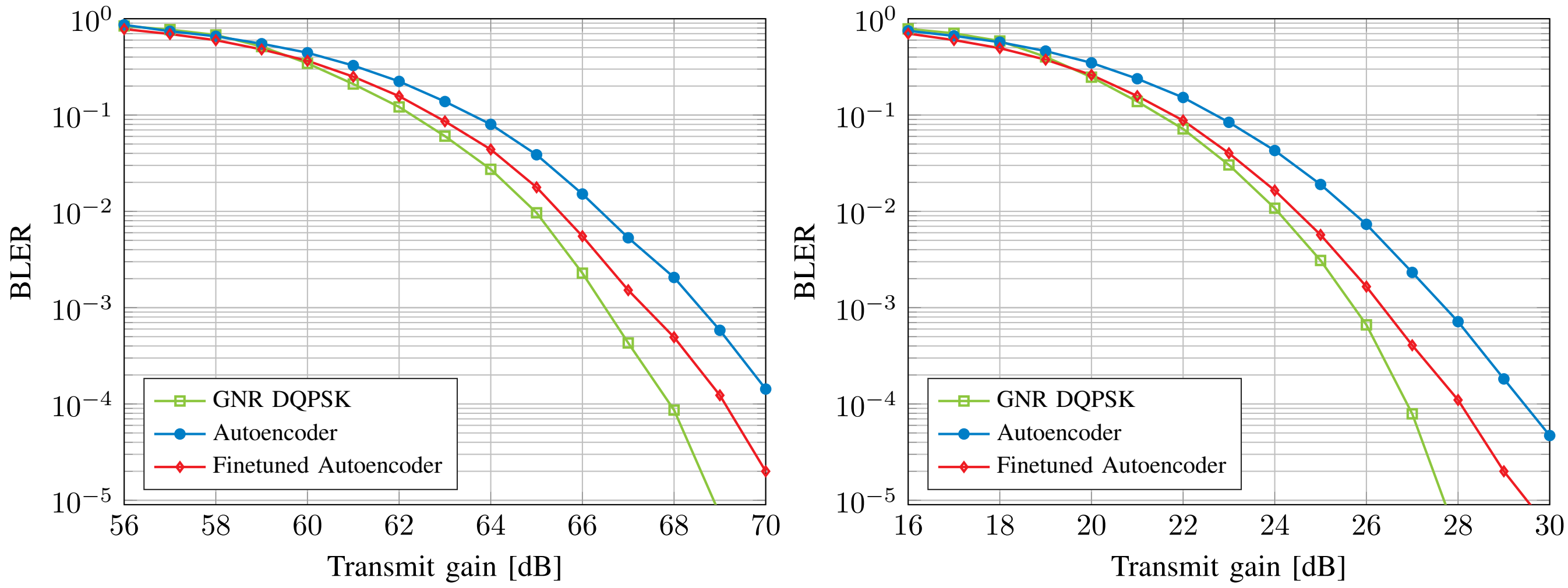

(a) Over-the-air

(b) Coaxial cable

Fig. 11: Measured BLER of the autoencoder (with and without fintuning) and the GNR DQPSK baseline